\documentclass{article}
\usepackage{ijcai24}

\usepackage{colortbl}  
\usepackage{xcolor}    
\usepackage{times}
\usepackage{soul}
\usepackage{url}
\usepackage[hidelinks]{hyperref}
\usepackage[utf8]{inputenc}
\usepackage[small]{caption}
\usepackage{graphicx}
\usepackage{amsmath}
\usepackage{amsthm}
\usepackage{booktabs}
\usepackage{algorithm}
\usepackage{algorithmic}
\usepackage[switch]{lineno}
\usepackage{bbm}
\usepackage{amssymb}
\usepackage{multirow}
\usepackage{subfigure}

\title{FedIDM: Achieving Fast and Stable Convergence in Byzantine Federated Learning through Iterative Distribution Matching}

\author{
He Yang
\and
Dongyi Lv\and
Wei Xi\and
Song Ma\and
Hanlin Gu\and
Jizhong Zhao\\
}

\begin{document}

\maketitle

\begin{abstract}
    Most existing Byzantine-robust federated learning (FL) methods suffer from slow and unstable convergence. Moreover, when handling a substantial proportion of colluded malicious clients, achieving robustness typically entails compromising model utility. To address these issues, this work introduces FedIDM, which employs distribution matching to construct trustworthy condensed data for identifying and filtering abnormal clients. FedIDM consists of two main components: (1) attack-tolerant condensed data generation, and (2) robust aggregation with negative contribution-based rejection. These components exclude local updates that (1) deviate from the update direction derived from condensed data, or (2) cause a significant loss on the condensed dataset. Comprehensive evaluations on three benchmark datasets demonstrate that FedIDM achieves fast and stable convergence while maintaining acceptable model utility, under multiple state-of-the-art Byzantine attacks involving a large number of malicious clients.
\end{abstract}

\section{Introduction}

Federated learning (FL)~\cite{kairouz2021advances,huang2024federated,yang2019federated,lyu2022privacy} facilitates learning from decentralized data sources while preserving privacy, enabling a wide range of promising, privacy-enhancing applications. Despite its advantages, the non-transparent
nature of local training data and processes renders FL vulnerable to various Byzantine attacks~\cite{wan2023four,li2023experimental,dong2023privacy,zhao2024huber,liu2024badsampler}, where the attacker-controlled malicious clients corrupt the model utility by submitting manipulated model updates.
\begin{figure}
\centering
\includegraphics[scale=0.86]{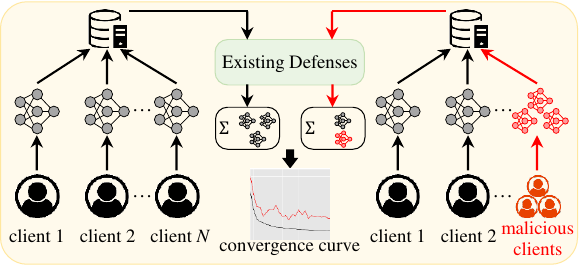}
\caption{Problem illustration. When handling a substantial proportion of colluded malicious clients, most existing Byzantine-robust methods may inadvertently discard valuable updates while preserving poisoned ones, hindering fast and stable convergence.}
\label{fig:illustration}
\vspace{-0.3cm}
\end{figure}


Thus far, considerable efforts have been dedicated to developing Byzantine-robust methods, which can be broadly classified into two categories. Firstly, one prevalent approach focuses on eliminating geometrical outliers using statistical techniques such as ~\cite{blanchard2017machine,guerraoui2018hidden,yin2018byzantine}. Secondly, another line of research explores anomaly detection techniques, including consistency checks~\cite{zhang2022fldetector}, update inversion~\cite{zhao2022fedinv}, and trust bootstrapping~\cite{cao2021fltrust}, to mitigate the impact of poisoned updates. Although these defenses demonstrate effectiveness against Byzantine attacks in certain scenarios, they face an important limitation when a substantial proportion of malicious clients engage in collusion. In such a case, these methods may inadvertently discard valuable updates from honest participants while retaining poisoned ones, hindering \textit{\textbf{fast and stable convergence}} (see Figure~\ref{fig:illustration}), which is crucial for the effectiveness of Byzantine-robust FL methods~\cite{huang2024federated}.

\begin{figure*}
\centering
\includegraphics[scale=1]{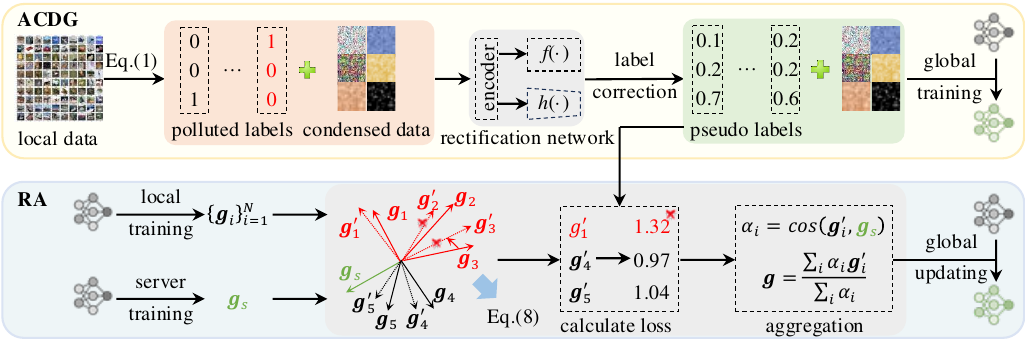}
\caption{Framework of FedIDM. $f(\cdot)$ and $h(\cdot)$ denote the feature extractor and classifier, respectively.}
\label{fig:defense}
\end{figure*}

To address this issue, we introduce FedIDM, an efficient Byzantine-robust federated learning framework based on iterative distribution matching(in Figure~\ref{fig:defense}). FedIDM divides the federated training process into two stages. In the former stage, FedIDM employs iterative distribution matching to construct compact synthetic datasets (\textit{i.e.}, condensed data) for global model training, which encapsulates rich and representative information, thereby accelerating convergence. In the latter stage, the server adjusts each client's local update using historical information and evaluates them on the condensed data. These updates are then aggregated using robust aggregation (\textbf{RA}) with a negative contribution-based rejection strategy to mitigate the impact of poisoned updates, preserving model utility and ensuring stable convergence.

However, a significant challenge in FedIDM arises when constructing trustworthy condensed data, as malicious clients may launch label-flipping attacks~\cite{jiang2023data,jebreel2024lfighter,jebreel2023fl} during this process. Defending against such attacks is challenging for two reasons: Firstly, the compression process in distribution matching necessitates a delicate balance between efficiency and information granularity, potentially causing the loss of critical details needed to detect subtle alterations. Furthermore, data heterogeneity introduces diversity in condensed data, complicating the establishment of a standardized anomaly detection mechanism. Therefore, we introduce an attack-tolerant condensed data generation (\textbf{ACDG}) scheme via contrastive label rectification into FedIDM. Concretely, contrastive learning combined with a Gaussian Mixture Model (GMM) is used to update a rectification network, which mitigates label pollution caused by label-flipping attacks in the condensed data. The rectification network processes each data point, generating model predictions as pseudo-labels. The pseudo-labeled condensed data are then utilized to update the global model.

Overall, our contributions can be summarized as follows:

\begin{itemize}
    \item 
    We introduce FedIDM, which leverages iterative distribution matching to ensure robustness while facilitating fast and stable convergence.
    \item 
    FedIDM has a minimal impact on model utility, even in scenarios involving a significant proportion of colluded malicious clients.
    \item 
    We empirically evaluate FedIDM against multiple state-of-the-art attacks on three benchmark datasets, demonstrating a substantial improvement compared to existing defenses.
\end{itemize}

\section{Related Work}
\paragraph{Byzantine Attacks against FL.}
In FL, Byzantine attacks are designed to diminish the global model's performance by corrupting a portion of the training data or manipulating local model updates during aggregation. In this study, we investigate three state-of-the-art Byzantine attacks: little is enough (LIE)~\cite{baruch2019little}, static optimization (STAT-OPT)~\cite{shejwalkar2021manipulating,shejwalkar2022back}, and Dynamic Optimization (DYN-OPT)~\cite{shejwalkar2021manipulating,shejwalkar2022back}. 

\textit{LIE}: The Little Is Enough (LIE) attack subtly corrupts the aggregation process by injecting minimal noise into each dimension of the averaged benign updates. Initially, the adversary calculates the average ($\nabla^b$) and standard deviation ($\sigma$) of the benign updates. Subsequently, a scaling coefficient ($z$) is derived from the ratio of compromised to benign clients. The adversarial update is formulated as $\nabla'=\nabla^b+z\cdot\sigma$. 

\textit{STAT-OPT}: The Static Optimization (STAT-OPT) attack introduces a general framework for poisoning in FL and customizes it for specific aggregation rules. The attack initiates by computing the average benign update, denoted as $\nabla^b$. It then determines a static malicious direction, denoted as $\omega = -sign(\nabla^b)$. The final poisoned update, $\nabla'$, is computed as $-\gamma\omega$, where $\gamma$ is a suboptimal value chosen to bypass the target aggregation rule. 

\textit{DYN-OPT}: The Dynamic Optimization (DYN-OPT) begins by calculating the average of the benign updates, denoted as $\nabla^b$. It then perturbs this average in a dynamic, data-dependent malicious direction, $\omega$, to compute the final poisoned update, $\nabla'=\nabla^b+\gamma\omega$. The parameter $\gamma$ is chosen to be the largest value that effectively circumvents the target aggregation rule.

\paragraph{Byzantine-Robust FL Methods.}
Considerable research efforts have been dedicated to developing robust FL methods to counter Byzantine attacks.These approaches generally fall into four categories: statistical defenses, anomaly detection defenses, trust-bootstrapping defenses, and client-side defenses. However, several challenges persist: (1) The large number of participants and the non-IID nature of client data complicate the detection of poisoned updates, as local updates inherently exhibit high variability. (2) colluded malicious clients can blend poisoned updates with legitimate ones. (3) The adversary can leverage its potential knowledge to craft stealthy attacks, further complicating the detection of poisoning attempts. Consequently, existing defense strategies often suffer from slow and unstable convergence. Furthermore, in scenarios with a significant proportion of compromised and colluded clients, achieving robustness typically comes at the expense of model utility. These limitations underscore the need for advanced and more effective defense mechanisms.

\section{Methodology}

\begin{table}
  \caption{Notations and Definitions.}
  \label{tab:notaions}
  \centering
  \begin{tabular}{cp{6.5cm}}
    \toprule
    Notation & Definition\\
    \midrule
    $f(\cdot)$ & Feature extractor of the rectification network.\\
    $h(\cdot)$ & Classifier of the rectification network.\\
    $h(s_i)_k$ & Probability of $s_i$ belonging to the $k$-th class.\\
    $\gamma_{ik}$ & Probability of the $i$-th data point belonging to the $k$-th class.\\
    $s_i^{(1)},s_i^{(2)}$ & The transformed data of $s_i$. \\
    $\textit{\textbf{g}}_i$ & Local update of the $i$-th client.\\
    $\textit{\textbf{g}}_s$ & Base update on the server-side.\\
    $w$ & Global model\\
    $\mathcal{S}^t$ & The condensed data during the $t$-th round.\\
    \bottomrule
  \end{tabular}
\end{table}

This section focuses on the detailed design of FedIDM. For clarity, the symbols used in this work and their corresponding explanations are provided in Table~\ref{tab:notaions}.

\subsection{Threat Model}
\paragraph{Adversary’s Objective.} 
The training objective of FedIDM is defined as the following optimization problem: $w^*={\arg\min}_{w\in\mathbbm{R}^d}\mathbbm{1}[t\leq \mathcal{T}]\frac{1}{|\mathcal{S}|}\sum_{i=1}^{|\mathcal{S}|}\ell (s_i,y_i;w)+\mathbbm{1}[t>\mathcal{T}]\frac{1}{N}\sum_{i=1}^Nf_i(w)$, where $w\in\mathbbm{R}^d$ represents the parameter to be optimized, $t$ denotes the communication round between clients and the server, and $\mathcal{T}$ is a constant. $\mathcal{S}$ represents the condensed data, $s_i$ denotes a single data point in $\mathcal{S}$, $y_i$ corresponds to the annotated label, which could potentially be a dirty label from an adversary. $N$ is the total number of clients, each of whom conducts local training with the local objective $f_i:\mathbbm{R}^p\rightarrow R$ based on its private dataset $\mathcal{D}_i$. In Byzantine FL, the adversary seeks to disrupt convergence by injecting malicious updates, thereby hindering global model optimization and diminishing its utility in real-world applications.

\paragraph{Adversary’s Capabilities.}
Following the attack setting in previous works~\cite{fang2020local,shejwalkar2021manipulating,shejwalkar2022back}, FedIDM adopts and extends certain assumptions regarding the capabilities of the adversary: (1) The adversary cannot access the training data of honest clients but has visibility into the local updates of honest clients. (2) The adversary can control multiple clients, referred to as malicious clients, under the assumption that their number may constitute up to 50\% of the total clients. (3) The adversary repetitively and dynamically executes attacks throughout the training process across multiple rounds. In this work, we randomly designate 50\% of the total rounds as adversarial rounds. (4) The adversary cannot influence the server (such as altering aggregation rules) or interfere with the training process of other honest clients.

\subsection{Attack-tolerant Condensed Data Generation}
During the attack-tolerant condensed data generation (ACDG) stage, clients locally generate a series of condensed data and then communicate condensed data to the central server. Subsequently, the server aggregates condensed data from all clients to facilitate the training of a global model. Inspired by recent advances in data distillation~\cite{zhao2023dataset,zhao2023improved,wang2022cafe,xiong2023feddm}, we formulate the following minimization problem to generate condensed data:

\begin{equation}
    \begin{aligned}
        \mathcal{S}^*&=\arg\min_{\mathcal{S}}\mathbbm{E}_{\theta\sim P_{\theta}}\|\frac{1}{|\mathcal{D}|}\sum_{i=1}^{|\mathcal{D}|}\phi_\theta(x_i)-\frac{1}{|\mathcal{S}|}\sum_{j=1}^{|\mathcal{S}|}\phi_\theta(s_j)\|^2\\
        &+\upsilon\mathcal{L}_{CE}(\mathcal{S})
    \end{aligned}
\end{equation}

where $\mathcal{D}$ and $\mathcal{S}$ denote the local training data and condensed data, respectively. $P_\theta$ represents the distribution of the randomly initialized network parameters, and $\phi_\theta$ refers to the model $\phi$ with parameters $\theta$. $\mathcal{L}_{CE}$ indicates the cross-entropy loss, and $\upsilon$ represents the regularization term. For simplicity, we omit the client subscript, as all clients follow the same optimization process.

However, this process is vulnerable to label-flipping attacks by the adversary, potentially disrupting model training. Therefore, we propose a contrastive label rectification method that incorporates a rectification network, consisting of a shared encoder, a feature extractor $f(\cdot)$, and a classifier $h(\cdot)$. ACDG involves two main stages: (1) modeling the distribution of condensed data. (2) generating pseudo-labels as the rectified labels and updating the global model.

\paragraph{Modeling the Distribution of Condensed Data.}
Firstly, ACDG utilizes a sliding window mechanism combined with semantic transformations for data augmentation. Specifically, at round $t$, we consolidate condensed data from $t-(\delta-1)$ to $t$ and feed them into the rectification network to cleanse potentially polluted labels. Given the aggregated condensed data $\Bar{\mathcal{S}}$ (defined as $\{\mathcal{S}^{t-(\delta-1)},\cdots,\mathcal{S}^t\}$), comprising $m$ data points $\{s_1,\cdots,s_m\}$, we apply semantic transformations through data augmentation techniques (such as flipping and cropping) to each data point twice. This yields a new set $\widetilde{\mathcal{S}}=\{s^{(1)}_1,s^{(2)}_1,\cdots,s^{(1)}_{m},s^{(2)}_m\}$, where $s^{(1)}_i$ and $s^{(2)}_i$ denote transformed data points.

Subsequently, a GMM is utilized to model the distribution of $\widetilde{\mathcal{S}}$ over its encoded representation $\textit{\textbf{r}}=f(\widetilde{\mathcal{S}})$. Upon introducing discrete latent variables $z\in\{1,2,\cdots,K\}$, which correspond to the distinct categories within the original dataset, the GMM is specified as follows:
\begin{equation}
    \begin{aligned}
        p(\textit{\textbf{r}})&=\sum_{k=1}^Kp(\textit{\textbf{r}},z=k)=\sum_{k=1}^Kp(z=k)\mathcal{N}(\textit{\textbf{r}};\mu_k,\sigma_k\mathbbm{I})
    \end{aligned}
\end{equation}

where $\mu_k$ and $\sigma_k$ represent the mean vector and standard deviation scalar of the $k$-th component in the GMM. $\mathbbm{I}$ denotes the identity matrix.

For the sake of convenience in description, we denote $\widetilde{\mathcal{S}}$ as $\{s'_1,\cdots,s'_{2m}\}$ here. During the parameter update phase of the GMM, we employ the model predictions $h(s'_i)$ for $i=1,\cdots,2m$ to guide the parameter adjustment process. Specifically, we integrate the model predictions into a standard Expectation-Maximization (EM) algorithm~\cite{dempster1977maximum}, replacing the traditional posterior probabilities. Consequently, the updated GMM parameters are determined as follows:
\begin{equation}
    \begin{aligned}
        \mu_k&=\frac{\sum_{i=1}^{2m}h(s'_i)_k\textit{\textbf{r}}_i}{\sum_ih(s'_i)_k}\\
        \sigma_k&=\frac{\sum_{i=1}^{2m}h(s'_i)_k(\textit{\textbf{r}}_i-\mu_k)(\textit{\textbf{r}}_i-\mu_k)^T}{\sum_ih(s'_i)_k}
    \end{aligned}
\end{equation}

where $h(s'_i)_k$ is the probability that the rectification network assigns data point $s'_i$ to the $k$-th category. Additionally, $\textit{\textbf{r}}_i$ denotes the embedded representation of $s'_i$.

\paragraph{Generating Pseudo-labels}
After constructing the GMM, we proceed to calculate posterior probabilities $\gamma_{ik}$, which quantify the probability that the $i$-th data point is assigned the $k$-th category:
\begin{equation}
    \gamma_{ik}=\frac{\exp\left(-(\textit{\textbf{r}}_i-\mu_k)^T(\textit{\textbf{r}}_i-\mu_k)/2\sigma_k\right)}{\sum_{k=1}^K\exp\left(-(\textit{\textbf{r}}_i-\mu_k)^T(\textit{\textbf{r}}_i-\mu_k)/2\sigma_k\right)}
\end{equation}

For each original sample $s_i$ with its claimed label $y_i$, we determine the probability of the sample belonging to category $k=y_i$ based on $\gamma_{ik}$. After calculating the posterior probabilities for the condensed data, we construct a two-component GMM, defined as follows:
\begin{equation}
    p(\gamma_{iy_i})=\sum_{b=0}^1p(\gamma_{iy_i},b)
\end{equation}

where $b$ is a binary latent variable representing the probability that $\gamma_{iy_i}$ belongs to a benign or malicious component. Subsequently, FedIDM relabels the transformed data points within $\widetilde{\mathcal{S}}$:
\begin{equation}
\left\{
    \begin{aligned}
        \tilde{y}^{(1)}_i&=\beta_iy_i+(1-\beta_i)h(s^{(1)}_i)\\
        \tilde{y}^{(2)}_i&=\beta_iy_i+(1-\beta_i)h(s^{(2)}_i)
    \end{aligned}
\right.
\end{equation}

where $\beta_i\in[0,1]$ represents the posterior probability that $s_i$ is free from pollution, as estimated by the two-component GMM, and $y_i$ denotes the claimed label.

After relabeling the condensed knowledge, we calculate the loss for updating the rectification network:
\begin{equation}
    \begin{aligned}
        \mathcal{L}=\mathcal{L}_{ce}+\mathcal{L}_{ctr}+\mathcal{L}_{mixup}
    \end{aligned}
\end{equation}

where $\mathcal{L}_{ce}$ denotes the cross-entropy loss, which is given by $\sum_{i=1}^m\left(\ell\left(h(s^{(1)}_i),\tilde{y}^{(2)}_i\right)+\ell\left(h(s^{(2)}_i),\tilde{y}^{(1)}_i\right)\right)$, ensuring that the rectification network's predictions for a pair of transformed data points are consistent. $\mathcal{L}_{ctr}$ is the InfoNCE loss~\cite{oord2018representation} defined as $\sum_{s^{(1)}_i,s^{(2)}_i\in\tilde{\mathcal{S}}}-\log\frac{\exp\left(f(s^{(1)}_i)^Tf(s^{(2)}_i)/\tau\right)}{\sum_{s_j\in\Bar{\mathcal{S}}\backslash s_i}\exp\left(f(s^{(1)}_i)^Tf(s_j)/\tau\right)}$, which promotes the learning of meaningful representations. Furthermore, $\mathcal{L}_{mixup}$ is the cross-entropy loss for a mixed sample using the Mixup method~\cite{zhang2018mixup}, calculated as $\ell(h(s^{(m)}_i), \tilde{y}^{m}_i)$, where $s^{(m)}_i=\rho s_i+(1-\rho)s_j,\tilde{y}^{(m)}_i=\rho((\tilde{y}^{(1)}_i+\tilde{y}^{(2)}_i)/2)+(1-\rho)((\tilde{y}^{(1)}_j+\tilde{y}^{(2)}_j)/2)$, and $s_i,s_j\in\Bar{\mathcal{S}}$.

When the rectification network is updated, FedIDM utilizes it to generate pseudo-labels for the aggregated condensed knowledge $\Bar{\mathcal{S}}$. For each $s_i\in\Bar{\mathcal{S}}$, the pseudo-label $\tilde{y}_i$ is computed using $h(s_i)$. Then these pseudo-labeled samples are integrated into the global model to facilitate its update.

\subsection{Robust Aggregation with Negative Contribution-based Rejection}
During the Robust Aggregation(RA) stage, each client independently refines a replica of the local model using its local data and then transmits the local update to the central server. During this process, the adversary may potentially execute attacks, which can compromise the integrity of the global model. In parallel, the server trains a separate replica of the global model using condensed knowledge, generating a base update. Subsequently, the server evaluates the contribution of each local update by calculating the cosine distance to the base update. This metric quantifies the alignment between the local and global updates, enabling the identification of updates with negative contributions. Local updates with negative contributions are discarded, while those with positive contributions are aggregated through a weighted sum based on their contributions.

\begin{figure*}[h]
  \setlength{\abovecaptionskip}{5.6pt}
  \centering
  \begin{subfigure}
    \centering
    \includegraphics[width=1.0\textwidth,keepaspectratio]{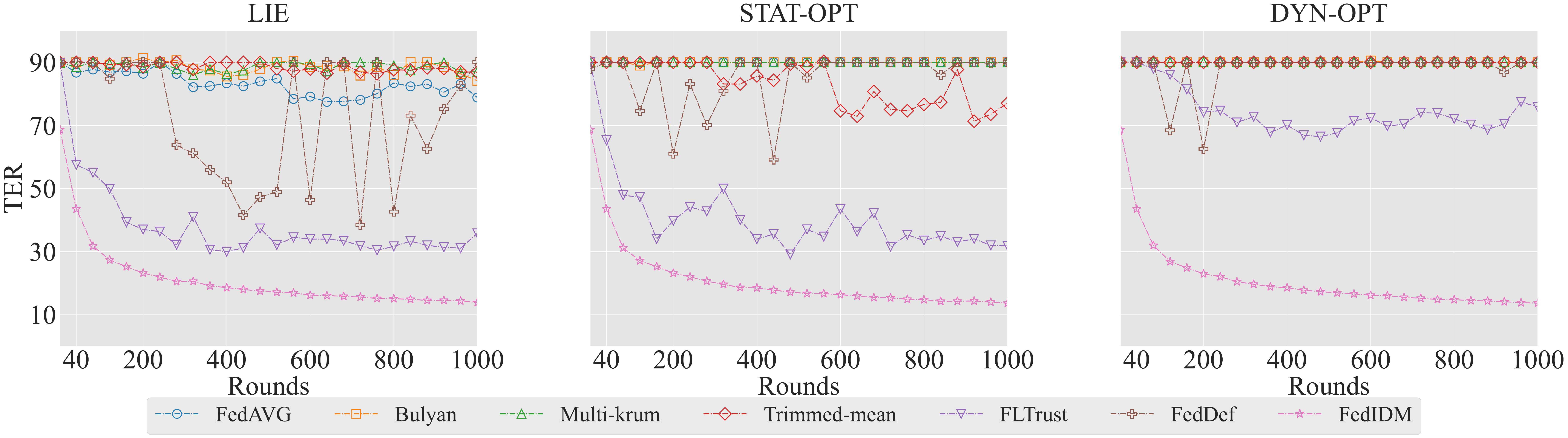}
    \caption{The convergence results on the CIFAR-10 dataset.}
    \label{fig:cifar10_convergence}
  \end{subfigure}
  \begin{subfigure}
    \centering
    \includegraphics[width=1.0\textwidth,keepaspectratio]{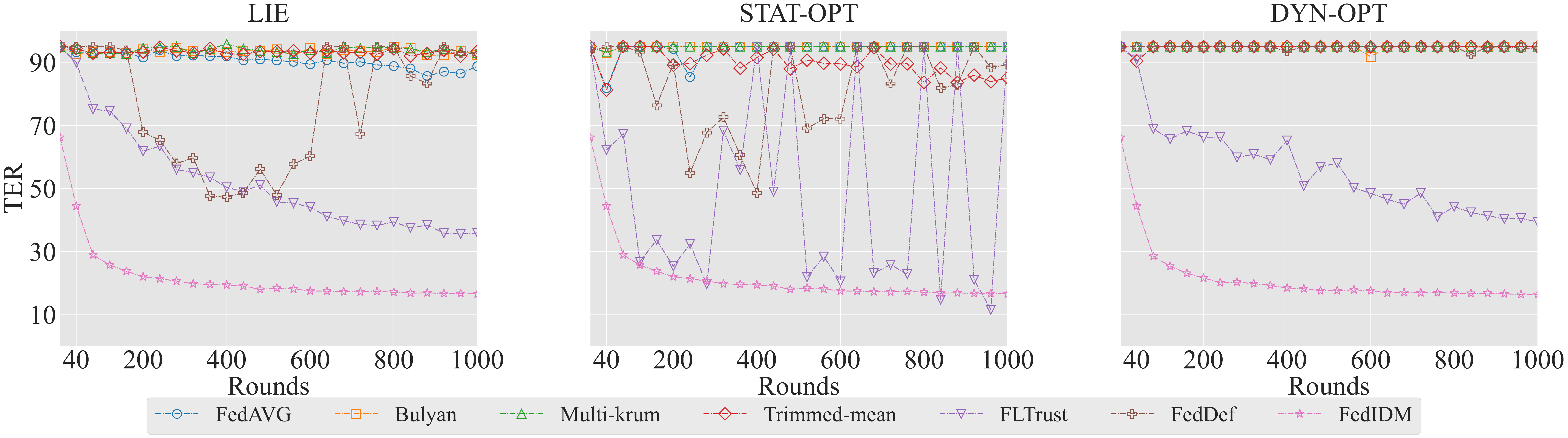}
    \caption{The convergence results on the CIFAR-100 dataset.}
    \label{fig:cifar100_convergence}
  \end{subfigure}
\end{figure*}

\paragraph{Contribution Evaluation.} 
In the $t$-th round of the interaction between clients and the server, each client $i$ computes a local update $\textit{\textbf{g}}^t_i$ based on its private data. Simultaneously, the server maintains a memory pool, denoted as $G^t=\{\textit{\textbf{g}}^{t-\Delta},\cdots,\textit{\textbf{g}}^{t-1}\}$, storing historical global updates from the recent $\Delta$ rounds, where $\textit{\textbf{g}}^t$ represents the global update at round $t$. Additionally, the server calculates a base update $\textit{\textbf{g}}^t_s$ using the aggregated knowledge $\Bar{\mathcal{S}}$ with pseudo-labels.

Firstly, a local update correction is conducted to exclude those updates that significantly differ from the server's base update, $\textit{\textbf{g}}^t_s$. To achieve this, we recalibrate each local update using a moving average between historical global updates and the local update. This correction aims to ensure the stationarity of the parameter updating process. Then a contribution $\alpha_i$ of $\textit{\textbf{g}}^t_i$ is evaluated based on the cosine similarity between this average and $\textit{\textbf{g}}^t_s$, determining whether to discard $\textit{\textbf{g}}^t_i$. If $\alpha_i>0$, it indicates that $\textit{\textbf{g}}^t_i$ will be retained; otherwise, it will be discarded. 
\begin{equation}
    \begin{aligned}
    \textit{\textbf{g}}'^t_i&=\sum_{j=1}^\Delta\lambda^{j+1}\cdot\text{norm}(\textit{\textbf{g}}^{t-j})+\lambda^j\cdot\text{norm}\left(\textit{\textbf{g}}^t_i\right)\\
    \alpha_i&=\cos{\left(\textit{\textbf{g}}'^t_i,\textit{\textbf{g}}^t_s\right)}
    \end{aligned}
\end{equation}

where $\lambda$ is a constant, $\alpha_i$ denotes the contribution of client $i$, and norm$(\cdot)$ is $\ell_2$-normalization such that $\|\textit{\textbf{g}}^t_i\|$ is equal to 1. We primarily focus on the directional deviation of the $\textit{\textbf{g}}^t_i$ from the base update $\textit{\textbf{g}}^t_s$, which necessitates the exclusion of the magnitude dimension to enhance the reliability of the contribution evaluation. Consequently, FedIDM applies normalization to each local update and historical global updates. 

\paragraph{Anomaly Detection and Aggregation} Assuming that the recalibrated local updates having positive contribution constitute a plausible honest set $\chi'_t=\{\textit{\textbf{g}}'^t_i|\alpha_i>0\}$. Additionally, the corresponding set of original gradients is denoted as $\chi_t=\{\textit{\textbf{g}}^t_i|\alpha_i>0\}$.  Although they exhibit positive contributions, stealthy poisoned updates may still be present. Therefore, additional anomaly detection is required. 

Firstly, we apply the DBSCAN algorithm to cluster updates in $\chi'_t$ according to their contributions. The minimum number of samples required to form a cluster is set to 1, allowing even a single update to be considered a separate cluster. Within each cluster, one update is randomly selected to represent the entire cluster, as updates with similar contributions may have comparable effects. This process aims to reduce redundancy and mitigate the undue influence of anomalous updates on the global model.

Secondly, to prevent extreme deviations in the magnitude of local updates, we adjust the magnitude of each update in $\chi'_t$ to the median of the original local updates in $\chi_t$. This adjustment is formulated as follows:
\begin{equation}
    \begin{aligned}
        \textit{\textbf{g}}'^t_i&=\mathfrak{g}\cdot\text{norm}(\textit{\textbf{g}}'^t_i),\quad\textit{\textbf{g}}'^t_i\in\chi'^t\\
        \mathfrak{g}&=\text{median}(\|\{{\textit{\textbf{g}}^t_i}\}_{i=1}^{|\chi^t|}\|),\quad\textit{\textbf{g}}^t_i\in\chi^t
    \end{aligned}
\end{equation}

This step aims to address optimization issues caused by extreme updates from anomalous clients, thereby improving the stability of model training.

Furthermore, we filter effective updates by assessing their impact on global model optimization. Specifically, each update is applied to the global model, and the loss on the distilled data is computed. The top $K$ updates with the highest losses are then eliminated, and the set $\chi'_t$ changes to $\chi'_t=\{\textit{\textbf{g}}'^t_i|\alpha_i>0, \ell_{i}<\ell_{o}\}$, where $\ell_o$ is a fixed lower bound loss value. This process removes updates that are ineffective or detrimental to global model optimization, ensuring a more reliable optimization process.

Finally, the global update is computed by averaging the remained recalibrated local updates in $\chi'_t$, with each update weighted according to its contribution $\alpha_i$. Formally, the computation is expressed as follows:
\begin{equation}
    \textit{\textbf{g}}^t=\frac{1}{\sum_{\textit{\textbf{g}}'^t_i\in\chi'^t}\alpha_i}\sum_{\textit{\textbf{g}}'^t_i\in\chi'^t}\alpha_i\textit{\textbf{g}}'^t_i
\end{equation}

Subsequently, the server proceeds to update the global model using the resulting global update, as described below, where $\eta$ is the global learning rate.
\begin{equation}
    w^{t+1}\xleftarrow{}w^t-\eta\textit{\textbf{g}}^t
\end{equation}

\begin{figure*}[h]
  \centering
  \includegraphics[width=1.0\textwidth]{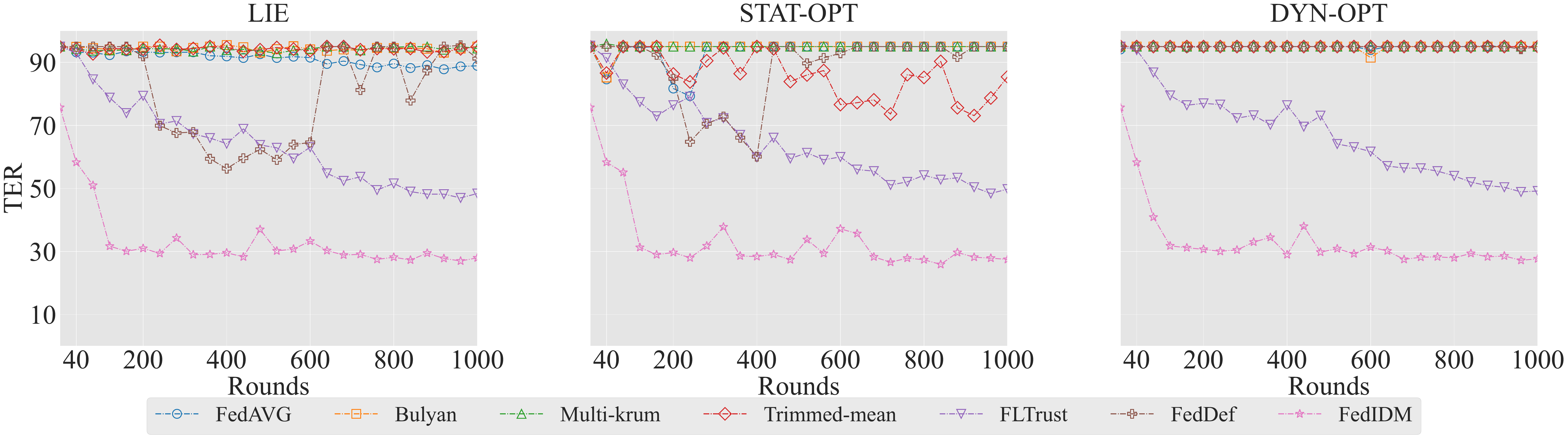}
  \caption{The convergence results on the Tiny-20 dataset.}
  \label{fig:tiny20_convergence}
\end{figure*}

\begin{table*}
    \renewcommand{\arraystretch}{1.05}
    \newcommand{\smallpm}[1]{\textcolor[HTML]{6AA84F}{\footnotesize $\pm$#1}}
    \caption{TERs of multiple FL defenses against Byzantine attacks. All values are reported as percentages.}
    \label{tab:FLDefenses}
    \centering
    \begin{tabular}{@{\hspace{0.1cm}}c@{\hspace{0.15cm}}c@{\hspace{0.15cm}}ccccccc}
        \toprule
        Dataset  & Attack & FedAVG & Bulyan & Multi-Krum & Trimmed-mean & FLTrust & FedDef & \cellcolor[HTML]{FFF2CC}FedIDM \\
        \midrule
        \multirow{3}{*}{CIFAR-10} 
     & LIE & 82.01\smallpm{7.40} & 87.83\smallpm{3.65} & 88.04\smallpm{1.96} & 88.71\smallpm{2.49} & 36.88\smallpm{5.02} & 84.60\smallpm{6.06} & \cellcolor[HTML]{FFF2CC}\textbf{14.08}\smallpm{0.29}\\
     & STAT & 90.00\smallpm{0.00} & 90.00\smallpm{0.00} & 90.00\smallpm{0.00} & 74.38\smallpm{8.14} & 34.01\smallpm{13.16} & 89.93\smallpm{0.26} & \cellcolor[HTML]{FFF2CC}\textbf{13.87}\smallpm{0.29}\\
     & DYN & 90.00\smallpm{0.00} & 90.00\smallpm{0.00} & 90.00\smallpm{0.00} & 90.00\smallpm{0.00} & 76.18\smallpm{0.67} & 89.97\smallpm{0.10} & \cellcolor[HTML]{FFF2CC}\textbf{13.75}\smallpm{0.19}\\
        \midrule

    \multirow{3}{*}{CIFAR-100} 
     & LIE & 87.94\smallpm{3.31} & 92.99\smallpm{2.26} & 92.98\smallpm{1.62} & 93.44\smallpm{1.56} & 35.74\smallpm{0.81} & 92.31\smallpm{6.61} & \cellcolor[HTML]{FFF2CC}\textbf{16.68}\smallpm{0.23}\\
     & STAT & 95.00\smallpm{0.00} & 95.00\smallpm{0.00} & 95.00\smallpm{0.00} & 85.95\smallpm{8.75} & 39.82\smallpm{1.08} & 91.52\smallpm{3.48} & \cellcolor[HTML]{FFF2CC}\textbf{16.32}\smallpm{0.23} \\
     & DYN & 95.00\smallpm{0.00} & 95.00\smallpm{0.00} & 95.00\smallpm{0.00} & 95.00\smallpm{0.00} & 39.50\smallpm{0.90} & 94.51\smallpm{0.96} & \cellcolor[HTML]{FFF2CC}\textbf{16.40}\smallpm{0.23} \\
        \midrule
    \multirow{3}{*}{Tiny-20} 
     & LIE & 89.05\smallpm{1.45} & 94.53\smallpm{1.23} & 94.34\smallpm{0.94} & 94.53\smallpm{1.03} & 46.80\smallpm{2.50} & 94.26\smallpm{3.06} & \cellcolor[HTML]{FFF2CC}\textbf{27.43}\smallpm{0.87}\\
     & STAT & 95.00\smallpm{0.00} & 95.00\smallpm{0.00} & 95.00\smallpm{0.00} & 82.96\smallpm{6.36} & 49.44\smallpm{1.24} & 95.00\smallpm{0.00} & \cellcolor[HTML]{FFF2CC}\textbf{27.98}\smallpm{0.68}\\
     & DYN & 95.00\smallpm{0.00} & 95.00\smallpm{0.00} & 95.00\smallpm{0.00} & 95.00\smallpm{0.00} & 49.37\smallpm{0.83} & 94.51\smallpm{0.96} & \cellcolor[HTML]{FFF2CC}\textbf{27.49}\smallpm{1.01} \\
    \bottomrule
    \end{tabular}
\end{table*}

\section{Experiment}
\paragraph{Datasets:}
We present experimental evaluations on three benchmark datasets for image classification: CIFAR-10, CIFAR-100~\cite{krizhevsky2009learning}, and Tiny-20~\cite{le2015tiny}. The CIFAR-10 and CIFAR-100 datasets consist of 50,000 $32\times32$ training images, distributed across 10 and 100 categories, respectively. For CIFAR-100, classification is performed using superclasses. The Tiny-20 dataset, derived from Tiny-ImageNet, is created by randomly selecting 20 classes and includes 10,000 $64\times64$ training images.

\paragraph{Federated Learning Setup and Attack Assumption:}
In our experiments, the default number of clients is set to $N=250$. During each communication round, 50 clients are randomly selected to contribute updates to the global model, which is constructed using a ResNet-18 or ResNet-34 architecture~\cite{he2016deep}. Consistent with prior studies~\cite{wang2020federated,li2021model}, we assume a Non-IID data distribution and utilize a Dirichlet distribution with a concentration parameter of 0.5 to partition data among clients. For experiments on CIFAR-100 and Tiny-20, the number of clients is set to $N=50$, with all clients participating in federated training during each round. The federated learning process spans 1,000 communication rounds. We randomly designate 50\% of the 1,000 rounds as adversarial rounds. During each adversarial round, we assume that the adversary compromises 50\% of the participating clients. During the ACDG stage of FedIDM, the adversary performs label-flipping attacks, including SLF~\cite{fang2020local} and DLF~\cite{shejwalkar2022back}, on the condensed data. During the RA stage in FedIDM, the adversary launches attacks including LIE, STAT-OPT, and DYN-OPT.

\paragraph{Evaluation Metrics:}
We employ the standard testing error rate (\textit{TER}) of the global model as the primary metric to assess the effectiveness of defense methods. A defense method is considered more robust if it yields lower TERs under adversarial conditions.

\paragraph{Convergence.}
Figure~\ref{fig:cifar10_convergence}, Figure~\ref{fig:cifar100_convergence}, and Figure~\ref{fig:tiny20_convergence} depict the convergence results of FedAVG~\cite{mcmahan2017communication}, Bulyan~\cite{guerraoui2018hidden}, Multi-Krum~\cite{blanchard2017machine}, Trimmed-mean~\cite{yin2018byzantine}, FLTrust~\cite{cao2021fltrust}, FedDef~\cite{park2023feddefender}, and FedIDM with respect to communication rounds. Statistical defenses are ineffective against potent attacks involving a significant proportion of compromised clients. FedIDM demonstrates superior robustness against various attacks. Furthermore, FedIDM achieves the fastest and most stable convergence compared to other methods. During the ACDG phase, clients efficiently extract representative condensed data, which accelerates model training compared to other FL approaches. In the subsequent RA phase, each local update is precisely evaluated and corrected, ensuring stable convergence throughout the training process.

\begin{figure*}
 \centering
 \subfigure[]{
  \begin{minipage}[t]{0.317\linewidth}
   \centering
   \includegraphics[width = 1\textwidth]{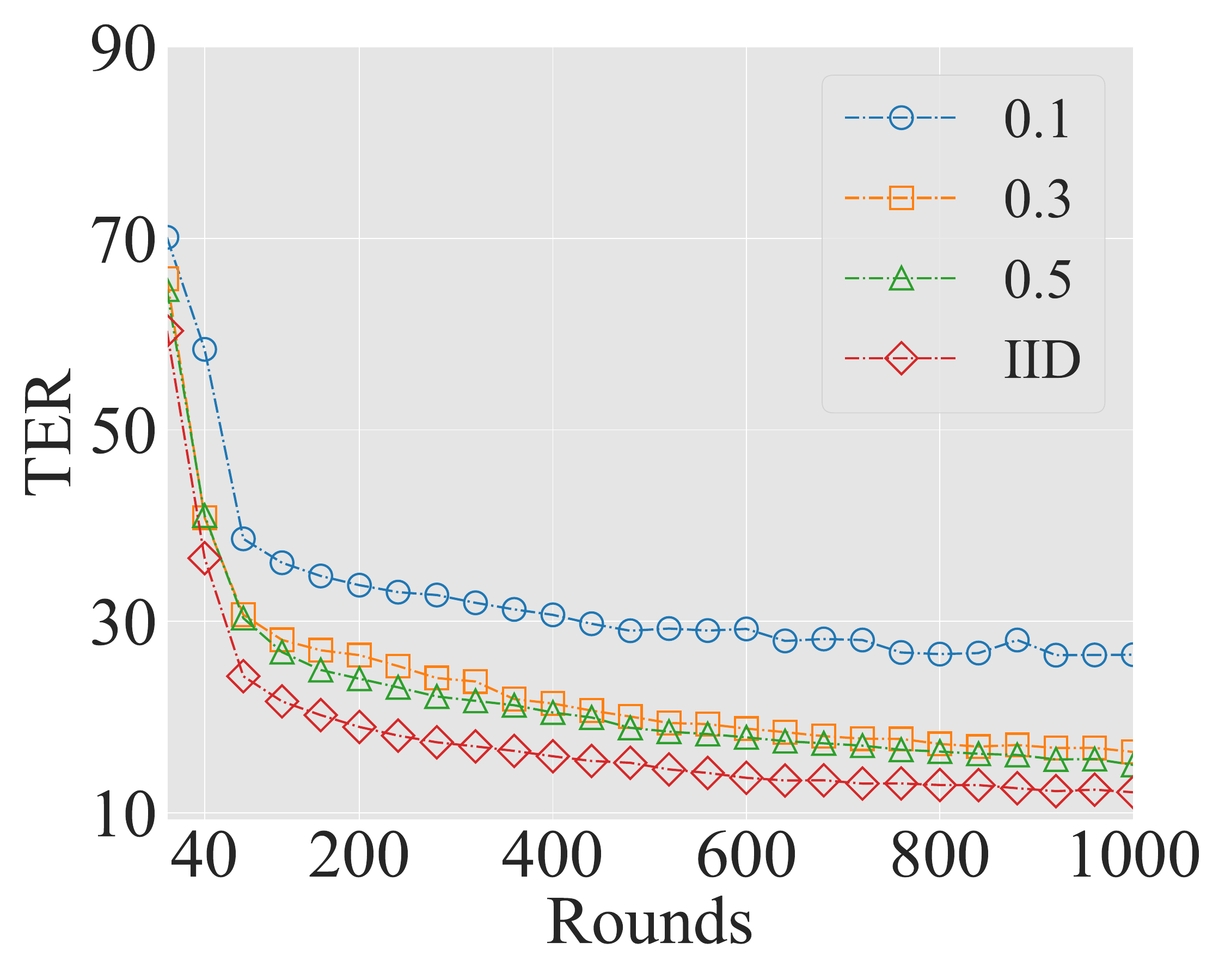}
  \end{minipage}
  \label{fig:noniid_extent}
 }
 \subfigure[]{
  \begin{minipage}[t]{0.317\linewidth}
   \centering
   \includegraphics[width = 1\textwidth]{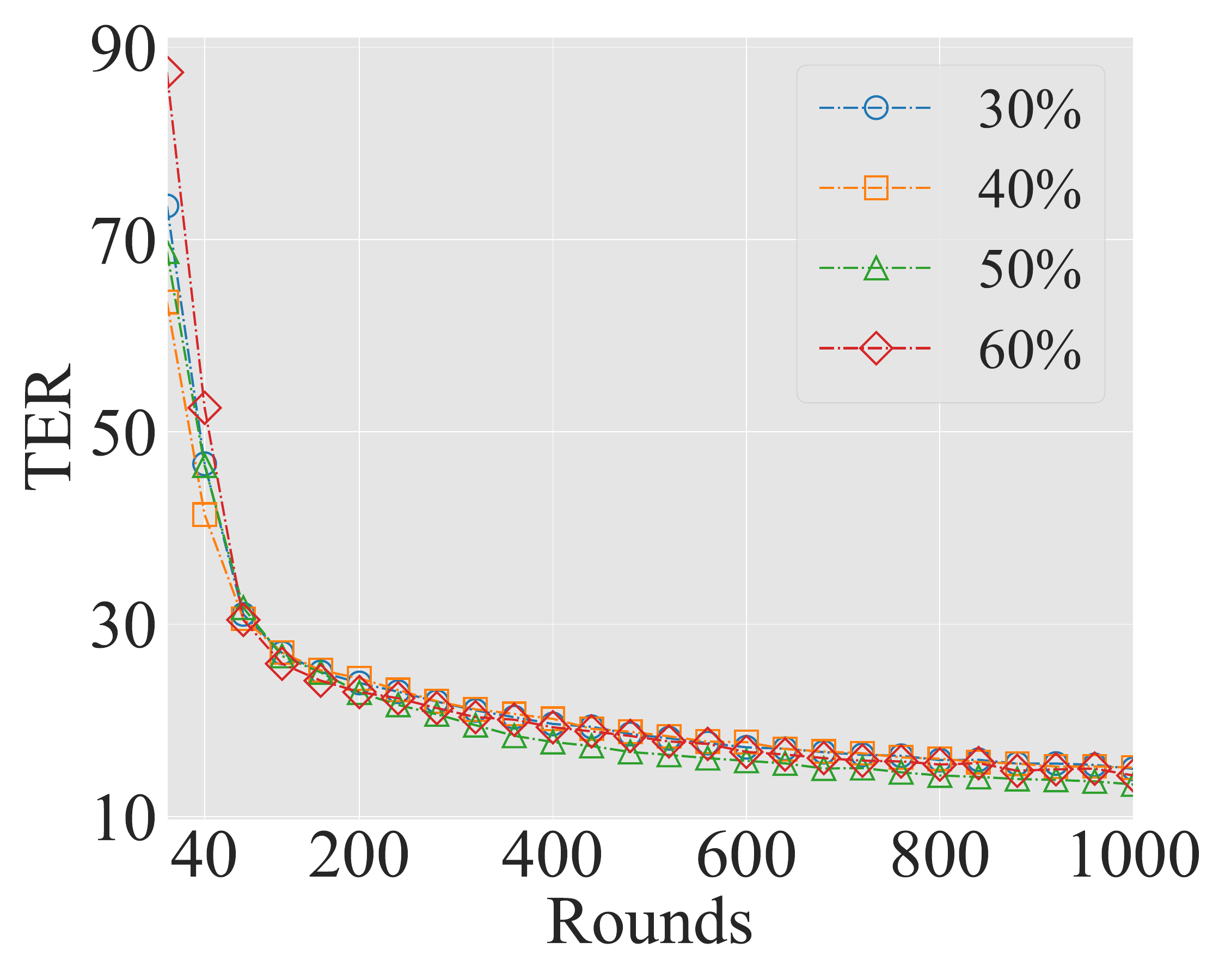}
  \end{minipage}
  \label{fig:attacker_ratio}
  }
  \subfigure[]{
  \begin{minipage}[t]{0.317\linewidth}
   \centering
   \includegraphics[width = 1\textwidth]{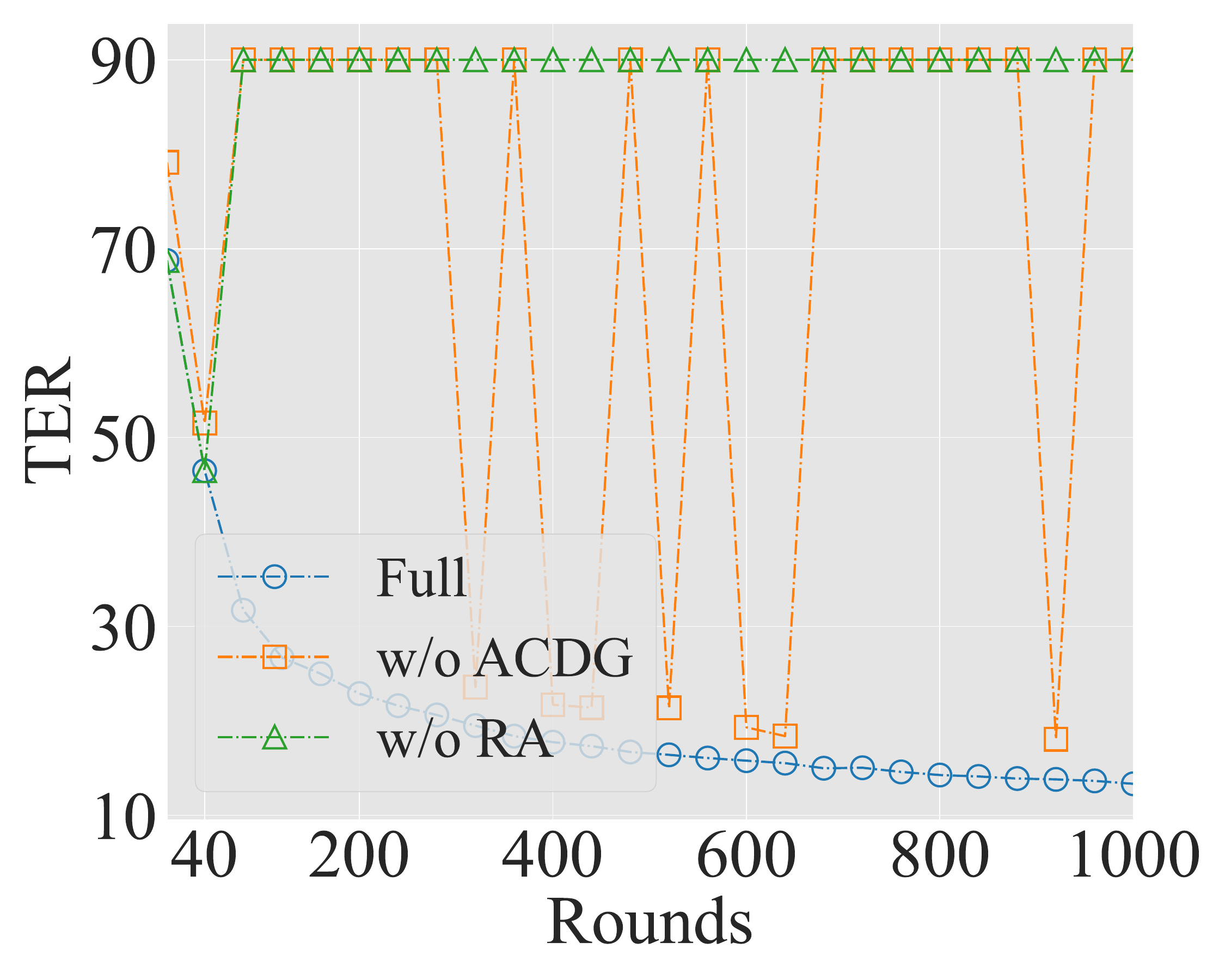}
  \end{minipage}
  \label{fig:ablation}
 }
 \caption{(a) Defense efficacy of FedIDM under varying degrees of Non-IID data distribution. (b) Defense efficacy of FedIDM on varying attacker ratios. (c) Effectiveness of different components within FedIDM, where ``full'' refers to the complete implementation of FedIDM.}
\end{figure*}

\paragraph{Defense Efficacy.}
As shown in Table~\ref{tab:FLDefenses}, FedIDM consistently achieves the lowest TERs across various datasets when defending against multiple attacks. Statistical defenses, such as Bulyan, Multi-Krum, and Trimmed-mean, fail to effectively filter anomalous updates due to the substantial proportion of compromised clients, leading to significant performance degradation. FLTrust, which assigns trust scores to local updates based on a root dataset, ensures robust aggregation. However, its performance can degrade if the root dataset deviates from the global data distribution, as shown in our experiments. Additionally, FedDefender, which introduces noise into the model, adversely impacts global model performance and fails to prevent Byzantine attacks. In contrast, FedIDM accurately evaluates the contribution of each local update and ensures a stable update process through anomaly detection, effectively mitigating the impact of poisoned updates and preserving model performance.

\paragraph{Defense against Non-IID extent.}
The extent of Non-IID data distribution significantly affects FedIDM, influencing both label rectification during the ACDG stage and the evaluation of local updates during the RA stage. As shown in Figure~\ref{fig:noniid_extent}, we assess the robustness of FedIDM under varying levels of Non-IID data distribution by adjusting the Dirichlet parameter. Specifically, we use Dirichlet distributions with parameters 0.1, 0.3, and 0.5 to simulate different degrees of Non-IID conditions. The results demonstrate FedIDM's effectiveness in mitigating Byzantine attacks across various Non-IID scenarios. However, as the degree of Non-IID data increases, a decline in model performance is observed. 

\paragraph{Defense Efficacy on Varying Attacker Ratio.}
We conducted additional experiments to examine the impact of varying proportions of compromised clients on the robustness of FedIDM. This variation provides deeper insights into the dynamics and resilience of FedIDM across different scenarios. Specifically, Figure~\ref{fig:attacker_ratio} presents results from experiments on the CIFAR-10 dataset, evaluating FedIDM's effectiveness against SLF and DYN-OPT attacks while varying the percentage of compromised clients. The findings demonstrate that FedIDM maintains resilience to these variations, underscoring its inherent robustness.

\paragraph{Effectiveness of Different Component.}
Finally, we conduct a comprehensive analysis of the individual components of FedIDM to evaluate their respective contributions. The ACDG stage plays a critical role in the effectiveness of FedIDM, as it is responsible for generating trustworthy condensed data, which is essential for subsequent robust aggregation. We introduce three auxiliary metrics: rectification success rate (RSR), false positive rate (FPR), and false negative rate (FNR). Specifically, RSR quantifies the proportion of corrupted labels that are accurately rectified. FPR reflects the fraction of legitimate labels that are incorrectly rectified, and FNR captures the percentage of polluted labels that are not properly rectified. Table~\ref{tab:effectiveness of ACDG} presents the RSRs, FPRs, and FNRs against SLF and DLF across different datasets. The results demonstrate the effectiveness of the ACDG stage in FedIDM, particularly in its ability to identify and rectify polluted labels while maintaining the integrity of legitimate ones. Furthermore, Figure~\ref{fig:ablation} provides an intuitive assessment of the effectiveness of each FedIDM component. Firstly, we deactivate the ACDG module, which causes FedIDM to fail against SLF and DYN-OPT attacks. Without ACDG, FedIDM cannot effectively mitigate label corruption, making it vulnerable to SLF attacks. Moreover, the absence of ACDG causes inaccurate evaluation of local update due to polluted condensed data, making FedIDM susceptible to DYN-OPT attacks. Subsequently, we deactivate the RA module and compute the average of the local updates, which again leads to insufficient defense against the DYN-OPT attack.

\begin{table}
  \caption{RSRs, FPRs, and FNRs of FedIDM against SLF and DLF.}
  \label{tab:effectiveness of ACDG}
  \centering
  \begin{tabular}{ccccc}
    \toprule
    Dataset & Attack & RSR & FPR & FNR\\
    \midrule

    \multirow{2}{*}{CIFAR-10} & SLF & 0.712 & 0.003 & 0.288 \\
     & DLF & 0.841 & 0.007 & 0.159 \\
     \midrule
     
    \multirow{2}{*}{CIFAR-100} & SLF & 0.763 & 0.037 & 0.237 \\
     & DLF & 0.971 & 0.004 & 0.029 \\
     \midrule

    \multirow{2}{*}{Tiny-20} & SLF & 0.757 & 0.037 & 0.243 \\
     & DLF & 0.927 & 0.016 & 0.072 \\
     \bottomrule
  \end{tabular}
\end{table}

\section{Conclusion}
In this study, we present FedIDM, a novel Byzantine-robust federated learning (FL) approach that ensures fast and stable convergence while addressing the challenges posed by colluded malicious clients. Specifically, FedIDM employs iterative distribution matching to generate reliable condensed data, facilitating the identification and filtering of abnormal clients. The method involves a two-stage process: (1) attack-tolerant condensed data generation, and (2) robust aggregation with negative contribution-based rejection. FedIDM effectively maintains model utility while ensuring robustness against a range of state-of-the-art Byzantine attacks. Extensive empirical evaluations on three benchmark datasets demonstrate that FedIDM outperforms existing defense mechanisms, effectively mitigating the impact of malicious clients without compromising model utility.







\bibliographystyle{named}
\bibliography{ijcai24}

@article{kairouz2021advances,
  title={Advances and open problems in federated learning},
  author={Kairouz, Peter and McMahan, H Brendan and Avent, Brendan and Bellet, Aur{\'e}lien and Bennis, Mehdi and Bhagoji, Arjun Nitin and Bonawitz, Kallista and Charles, Zachary and Cormode, Graham and Cummings, Rachel and others},
  journal={Foundations and trends{\textregistered} in machine learning},
  volume={14},
  number={1--2},
  pages={1--210},
  year={2021},
  publisher={Now Publishers, Inc.}
}

@article{huang2024federated,
  title={Federated learning for generalization, robustness, fairness: A survey and benchmark},
  author={Huang, Wenke and Ye, Mang and Shi, Zekun and Wan, Guancheng and Li, He and Du, Bo and Yang, Qiang},
  journal={IEEE Transactions on Pattern Analysis and Machine Intelligence},
  year={2024},
  publisher={IEEE}
}

@article{yang2019federated,
  title={Federated machine learning: Concept and applications},
  author={Yang, Qiang and Liu, Yang and Chen, Tianjian and Tong, Yongxin},
  journal={ACM Transactions on Intelligent Systems and Technology (TIST)},
  volume={10},
  number={2},
  pages={1--19},
  year={2019},
  publisher={ACM New York, NY, USA}
}

@article{lyu2022privacy,
  title={Privacy and robustness in federated learning: Attacks and defenses},
  author={Lyu, Lingjuan and Yu, Han and Ma, Xingjun and Chen, Chen and Sun, Lichao and Zhao, Jun and Yang, Qiang and Philip, S Yu},
  journal={IEEE transactions on neural networks and learning systems},
  year={2022},
  publisher={IEEE}
}

@inproceedings{wan2023four,
  title={A four-pronged defense against byzantine attacks in federated learning},
  author={Wan, Wei and Hu, Shengshan and Li, Minghui and Lu, Jianrong and Zhang, Longling and Zhang, Leo Yu and Jin, Hai},
  booktitle={Proceedings of the 31st ACM International Conference on Multimedia},
  pages={7394--7402},
  year={2023}
}

@article{li2023experimental,
  title={An experimental study of byzantine-robust aggregation schemes in federated learning},
  author={Li, Shenghui and Ngai, Edith C-H and Voigt, Thiemo},
  journal={IEEE Transactions on Big Data},
  year={2023},
  publisher={IEEE}
}

@article{dong2023privacy,
  title={Privacy-preserving and byzantine-robust federated learning},
  author={Dong, Caiqin and Weng, Jian and Li, Ming and Liu, Jia-Nan and Liu, Zhiquan and Cheng, Yudan and Yu, Shui},
  journal={IEEE Transactions on Dependable and Secure Computing},
  volume={21},
  number={2},
  pages={889--904},
  year={2023},
  publisher={IEEE}
}

@inproceedings{zhao2024huber,
  title={A huber loss minimization approach to byzantine robust federated learning},
  author={Zhao, Puning and Yu, Fei and Wan, Zhiguo},
  booktitle={Proceedings of the AAAI Conference on Artificial Intelligence},
  volume={38},
  number={19},
  pages={21806--21814},
  year={2024}
}

@inproceedings{liu2024badsampler,
  title={Badsampler: Harnessing the power of catastrophic forgetting to poison byzantine-robust federated learning},
  author={Liu, Yi and Wang, Cong and Yuan, Xingliang},
  booktitle={Proceedings of the 30th ACM SIGKDD Conference on Knowledge Discovery and Data Mining},
  pages={1944--1955},
  year={2024}
}

@article{blanchard2017machine,
  title={Machine learning with adversaries: Byzantine tolerant gradient descent},
  author={Blanchard, Peva and El Mhamdi, El Mahdi and Guerraoui, Rachid and Stainer, Julien},
  journal={Advances in neural information processing systems},
  volume={30},
  year={2017}
}

@inproceedings{guerraoui2018hidden,
  title={The hidden vulnerability of distributed learning in byzantium},
  author={Guerraoui, Rachid and Rouault, S{\'e}bastien and others},
  booktitle={International Conference on Machine Learning},
  pages={3521--3530},
  year={2018},
  organization={PMLR}
}

@inproceedings{yin2018byzantine,
  title={Byzantine-robust distributed learning: Towards optimal statistical rates},
  author={Yin, Dong and Chen, Yudong and Kannan, Ramchandran and Bartlett, Peter},
  booktitle={International Conference on Machine Learning},
  pages={5650--5659},
  year={2018},
  organization={Pmlr}
}

@inproceedings{zhang2022fldetector,
  title={Fldetector: Defending federated learning against model poisoning attacks via detecting malicious clients},
  author={Zhang, Zaixi and Cao, Xiaoyu and Jia, Jinyuan and Gong, Neil Zhenqiang},
  booktitle={Proceedings of the 28th ACM SIGKDD Conference on Knowledge Discovery and Data Mining},
  pages={2545--2555},
  year={2022}
}

@inproceedings{zhao2022fedinv,
  title={Fedinv: Byzantine-robust federated learning by inversing local model updates},
  author={Zhao, Bo and Sun, Peng and Wang, Tao and Jiang, Keyu},
  booktitle={Proceedings of the AAAI Conference on Artificial Intelligence},
  volume={36},
  number={8},
  pages={9171--9179},
  year={2022}
}

@inproceedings{cao2021fltrust,
  title={FLTrust: Byzantine-robust Federated Learning via Trust Bootstrapping},
  author={Cao, Xiaoyu and Fang, Minghong and Liu, Jia and Gong, Neil Zhenqiang},
  booktitle={ISOC Network and Distributed System Security Symposium (NDSS)},
  year={2021}
}

@inproceedings{fang2020local,
  title={Local model poisoning attacks to $\{$Byzantine-Robust$\}$ federated learning},
  author={Fang, Minghong and Cao, Xiaoyu and Jia, Jinyuan and Gong, Neil},
  booktitle={29th USENIX security symposium (USENIX Security 20)},
  pages={1605--1622},
  year={2020}
}

@inproceedings{park2023feddefender,
  title={Feddefender: Client-side attack-tolerant federated learning},
  author={Park, Sungwon and Han, Sungwon and Wu, Fangzhao and Kim, Sundong and Zhu, Bin and Xie, Xing and Cha, Meeyoung},
  booktitle={Proceedings of the 29th ACM SIGKDD conference on knowledge discovery and data mining},
  pages={1850--1861},
  year={2023}
}

@article{jiang2023data,
  title={Data quality detection mechanism against label flipping attacks in federated learning},
  author={Jiang, Yifeng and Zhang, Weiwen and Chen, Yanxi},
  journal={IEEE Transactions on Information Forensics and Security},
  volume={18},
  pages={1625--1637},
  year={2023},
  publisher={IEEE}
}

@article{jebreel2024lfighter,
  title={LFighter: Defending against the label-flipping attack in federated learning},
  author={Jebreel, Najeeb Moharram and Domingo-Ferrer, Josep and S{\'a}nchez, David and Blanco-Justicia, Alberto},
  journal={Neural Networks},
  volume={170},
  pages={111--126},
  year={2024},
  publisher={Elsevier}
}

@article{jebreel2023fl,
  title={Fl-defender: Combating targeted attacks in federated learning},
  author={Jebreel, Najeeb Moharram and Domingo-Ferrer, Josep},
  journal={Knowledge-Based Systems},
  volume={260},
  pages={110178},
  year={2023},
  publisher={Elsevier}
}

@article{baruch2019little,
  title={A little is enough: Circumventing defenses for distributed learning},
  author={Baruch, Gilad and Baruch, Moran and Goldberg, Yoav},
  journal={Advances in Neural Information Processing Systems},
  volume={32},
  year={2019}
}

@inproceedings{shejwalkar2021manipulating,
  title={Manipulating the byzantine: Optimizing model poisoning attacks and defenses for federated learning},
  author={Shejwalkar, Virat and Houmansadr, Amir},
  booktitle={NDSS},
  year={2021}
}

@inproceedings{shejwalkar2022back,
  title={Back to the drawing board: A critical evaluation of poisoning attacks on production federated learning},
  author={Shejwalkar, Virat and Houmansadr, Amir and Kairouz, Peter and Ramage, Daniel},
  booktitle={2022 IEEE Symposium on Security and Privacy (SP)},
  pages={1354--1371},
  year={2022},
  organization={IEEE}
}

@inproceedings{zhao2023dataset,
  title={Dataset condensation with distribution matching},
  author={Zhao, Bo and Bilen, Hakan},
  booktitle={Proceedings of the IEEE/CVF Winter Conference on Applications of Computer Vision},
  pages={6514--6523},
  year={2023}
}

@inproceedings{zhao2023improved,
  title={Improved distribution matching for dataset condensation},
  author={Zhao, Ganlong and Li, Guanbin and Qin, Yipeng and Yu, Yizhou},
  booktitle={Proceedings of the IEEE/CVF Conference on Computer Vision and Pattern Recognition},
  pages={7856--7865},
  year={2023}
}

@inproceedings{wang2022cafe,
  title={Cafe: Learning to condense dataset by aligning features},
  author={Wang, Kai and Zhao, Bo and Peng, Xiangyu and Zhu, Zheng and Yang, Shuo and Wang, Shuo and Huang, Guan and Bilen, Hakan and Wang, Xinchao and You, Yang},
  booktitle={Proceedings of the IEEE/CVF Conference on Computer Vision and Pattern Recognition},
  pages={12196--12205},
  year={2022}
}

@inproceedings{xiong2023feddm,
  title={Feddm: Iterative distribution matching for communication-efficient federated learning},
  author={Xiong, Yuanhao and Wang, Ruochen and Cheng, Minhao and Yu, Felix and Hsieh, Cho-Jui},
  booktitle={Proceedings of the IEEE/CVF Conference on Computer Vision and Pattern Recognition},
  pages={16323--16332},
  year={2023}
}

@article{dempster1977maximum,
  title={Maximum likelihood from incomplete data via the EM algorithm},
  author={Dempster, Arthur P and Laird, Nan M and Rubin, Donald B},
  journal={Journal of the royal statistical society: series B (methodological)},
  volume={39},
  number={1},
  pages={1--22},
  year={1977},
  publisher={Wiley Online Library}
}

@article{oord2018representation,
  title={Representation learning with contrastive predictive coding},
  author={Oord, Aaron van den and Li, Yazhe and Vinyals, Oriol},
  journal={arXiv preprint arXiv:1807.03748},
  year={2018}
}

@inproceedings{zhang2018mixup,
  title={mixup: Beyond Empirical Risk Minimization},
  author={Zhang, Hongyi and Cisse, Moustapha and Dauphin, Yann N and Lopez-Paz, David},
  booktitle={International Conference on Learning Representations},
  year={2018}
}

@article{krizhevsky2009learning,
  title={Learning multiple layers of features from tiny images},
  author={Krizhevsky, Alex and Hinton, Geoffrey and others},
  year={2009},
  publisher={Toronto, ON, Canada}
}

@article{le2015tiny,
  title={Tiny imagenet visual recognition challenge},
  author={Le, Ya and Yang, Xuan},
  journal={CS 231N},
  volume={7},
  number={7},
  pages={3},
  year={2015}
}

@inproceedings{he2016deep,
  title={Deep residual learning for image recognition},
  author={He, Kaiming and Zhang, Xiangyu and Ren, Shaoqing and Sun, Jian},
  booktitle={Proceedings of the IEEE conference on computer vision and pattern recognition},
  pages={770--778},
  year={2016}
}

@inproceedings{li2021model,
  title={Model-contrastive federated learning},
  author={Li, Qinbin and He, Bingsheng and Song, Dawn},
  booktitle={Proceedings of the IEEE/CVF conference on computer vision and pattern recognition},
  pages={10713--10722},
  year={2021}
}

@inproceedings{wang2020federated,
  title={Federated Learning with Matched Averaging},
  author={Wang, Hongyi and Yurochkin, Mikhail and Sun, Yuekai and Papailiopoulos, Dimitris and Khazaeni, Yasaman},
  booktitle={International Conference on Learning Representations},
  year={2020}
}

@inproceedings{mcmahan2017communication,
  title={Communication-efficient learning of deep networks from decentralized data},
  author={McMahan, Brendan and Moore, Eider and Ramage, Daniel and Hampson, Seth and y Arcas, Blaise Aguera},
  booktitle={Artificial intelligence and statistics},
  pages={1273--1282},
  year={2017},
  organization={PMLR}
}

\end{document}